\pdfoutput=1

\documentclass[11pt]{article}

\usepackage[final]{acl}

\usepackage{times}
\usepackage{latexsym}

\usepackage[T1]{fontenc}

\usepackage[utf8]{inputenc}

\usepackage{microtype}

\usepackage{inconsolata}
\usepackage{amssymb}
\usepackage{float}
\usepackage{booktabs}
\usepackage{multirow}
\usepackage{arydshln}
\usepackage{CJKutf8}
\usepackage{graphicx}
\newcommand{\ctslogo}{\raisebox{3.4pt}{\includegraphics[scale=0.0105]{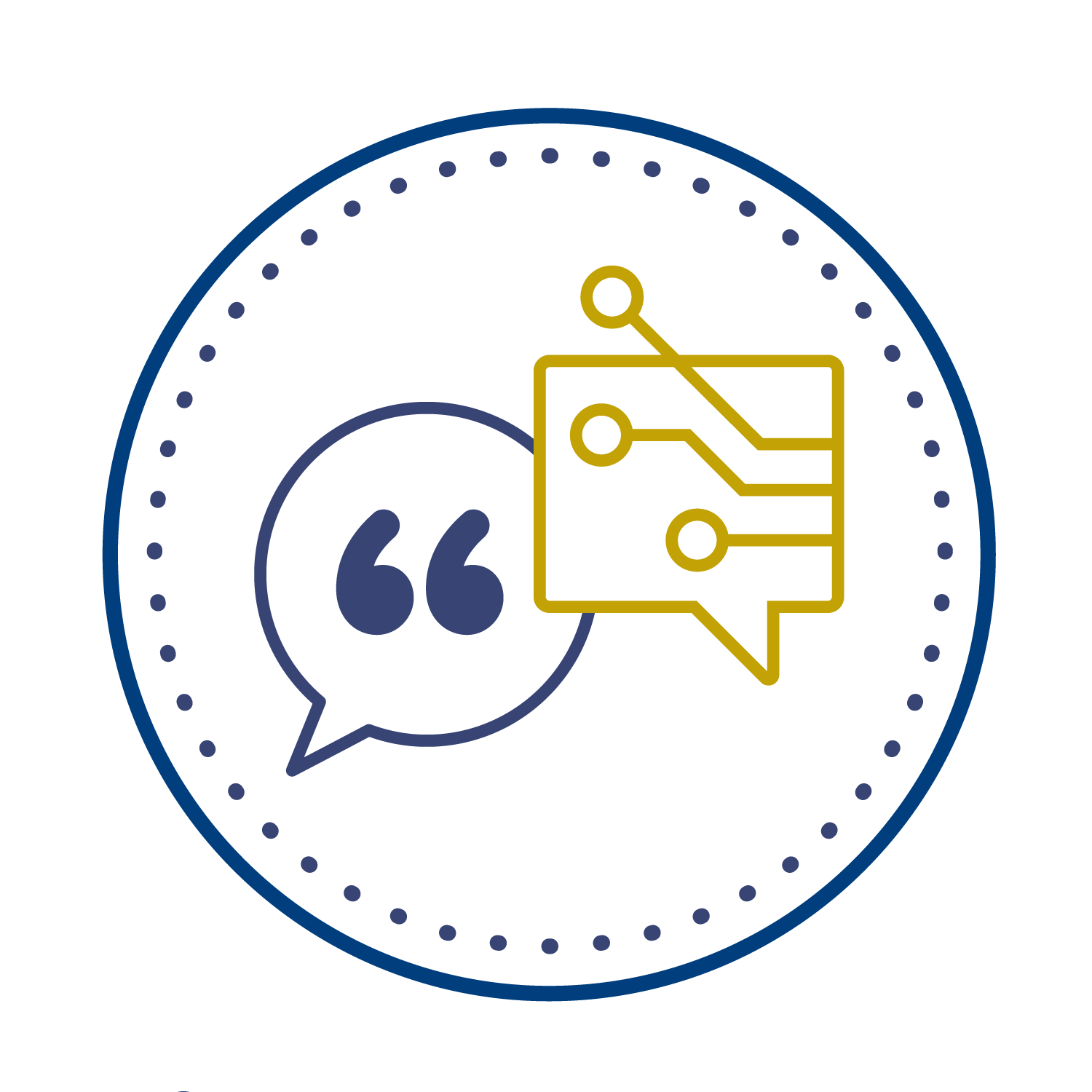}}}
\newcommand{\PAIlogo}{\raisebox{3.4pt}{\includegraphics[scale=0.083]{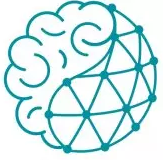}}}
\title{Are Large Language Models State-of-the-art Quality Estimators for Machine Translation of User-generated Content?}

\author{Shenbin Qian\ctslogo, Constantin Orăsan\ctslogo, Diptesh Kanojia\PAIlogo  and Félix do Carmo\ctslogo \\
\ctslogo Centre for Translation Studies and 
\PAIlogo Institute for People-Centred AI, \\[.2em]
University of Surrey, United Kingdom \\[.13em]
\{s.qian, c.orasan, d.kanojia, f.docarmo\}@surrey.ac.uk}

\begin{document}
\maketitle
\begin{abstract}
  This paper investigates whether large language models (LLMs) are state-of-the-art quality estimators for machine translation of user-generated content (UGC) that contains emotional expressions, without the use of reference translations. To achieve this, we employ an existing emotion-related dataset with human-annotated errors and calculate quality evaluation scores based on the Multi-dimensional Quality Metrics. We compare the accuracy of several LLMs with that of our fine-tuned baseline models, under in-context learning and parameter-efficient fine-tuning (PEFT) scenarios. We find that PEFT of LLMs leads to better performance in score prediction with human interpretable explanations than fine-tuned models. However, a manual analysis of LLM outputs reveals that they still have problems such as refusal to reply to a prompt and unstable output while evaluating machine translation of UGC. 
\end{abstract}

\section{Introduction}

\begin{figure*}[h]
  \centering
  \includegraphics[width=0.9\textwidth]{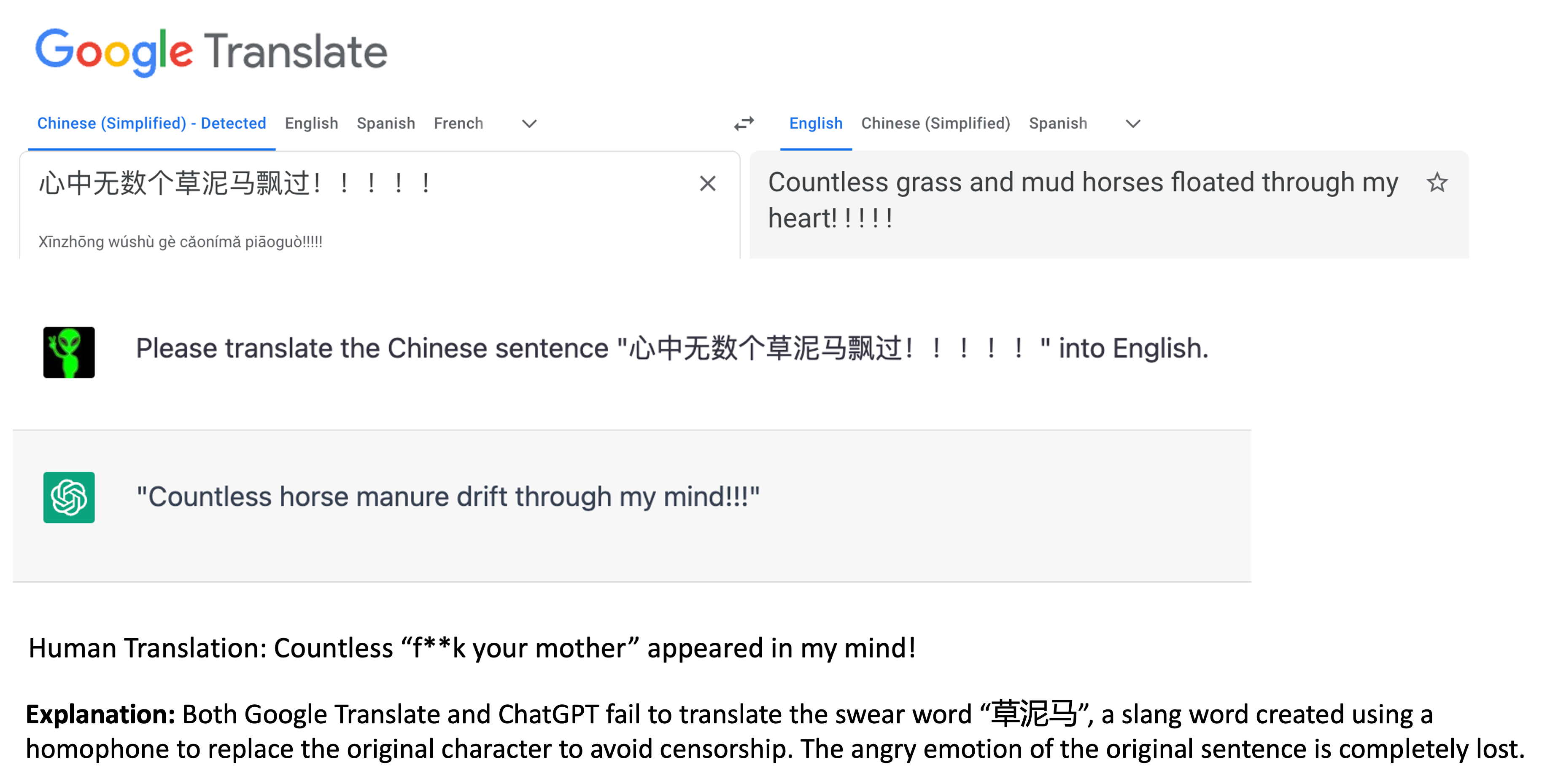}
  \caption{Example of translations from Google Translate and ChatGPT}
  \label{fig.example1}
\end{figure*}

Recent advancements in machine translation (MT) technology, particularly in Chinese-English news translation, have led to claims of achieving human parity~\citep{Hassan2018}. These claims have gained traction, particularly with the emergence of large language models (LLMs)~\citep{Wang2021-lh}, and their reported zero-shot state-of-the-art (SoTA) performance across various downstream tasks~\cite{OpenAI2023}. However, translating user-generated content (UGC) containing emotional expressions, such as tweets, poses additional challenges for MT systems~\citep{saadany-etal-2023-analysing}. As illustrated in Figure~\ref{fig.example1}, testing Google Translate (GT) and ChatGPT\footnote{GPT-3.5 at ``https://chat.openai.com/'' in Mar., 2024} using Chinese UGC with emotional slang revealed that the output of these systems requires significant improvement to be considered usable. This highlights the importance of evaluating MT quality using metrics that account for emotion preservation in translation.

Relying on human evaluation to assess the quality of machine translation is costly in terms of both time and money~\citep{Dorr2011,Lai2020}. Quality estimation (QE), which predicts MT quality in the absence of human references, can serve as a cost-effective alternative to approximate human evaluation~\citep{Specia2018}. A commonly adopted QE method involves fine-tuning multilingual pre-trained language models (PTLMs) on human evaluation data using frameworks like Multi-dimensional Quality Metrics (MQM), an error-based evaluation scheme for MT quality~\citep{Lommel2014}. These fine-tuned models can provide a score for MT outputs, indicating translation quality. However, this approach has faced criticism for its lack of explainability~\citep{Guerreiro2023-pf}. 

The inherent generative capability of LLMs allows for the provision of QE scores along with natural language explanations, rendering them comprehensible to humans. Some research claims that LLMs excel as quality evaluators in score prediction, in addition to their explainability~\citep{kocmi-federmann-2023-large}. Our paper delves into the question, \textit{``Are LLMs SoTA quality estimators for the translation of Chinese emotion-loaded UGC, through in-context learning (ICL)}\footnote{We refer to ICL as the ability of a LLM to adapt to new tasks by examples or instructions, without parameter updates or explicit training. It includes zero- and few-shot learning.} \textit{and parameter-efficient fine-tuning (PEFT)?''}. To answer this question, we utilize an existing dataset that was collected for the study of emotion translation in social media texts, and enhance it by adding segment-level QE scores based on MQM. This augmentation allows for the evaluation of LLMs' performance in predicting a QE score that reflects the overall translation quality of the MT segment. Our findings are contrasted with those of the conventional supervised fine-tuning approach. Our method achieves better results than fine-tuning on the emotion-related UGC dataset. Our contributions can be summarized as follows: 

\begin{itemize}
\itemsep 0mm

\item Computing QE score based on MQM for each data instance. 
\item Novel prompt templates for ICL and PEFT using multiple LLMs to evaluate MT quality of emotion-loaded UGC, achieving improved performance over the baseline with PEFT\footnote{\url{https://github.com/surrey-nlp/LLMs4MTQE-UGC}.}. 
\item Manually analyzing LLM outputs, and revealing problems such as \textit{refusal to reply} and \textit{unstable output}.

\end{itemize}

\section{Related Work} \label{lit}

Current state-of-the-art QE models are obtained by fine-tuning multilingual PTLMs on human evaluation data based on metrics such as translation edit rate (TER)~\citep{snover-etal-2006-study}, direct assessment (DA)~\citep{graham-etal-2013-continuous}, MQM and~\textit{etc}. For instance, TransQuest~\citep{Ranasinghe2020} employs the pre-trained XLM-RoBERTa~\citep{conneau-etal-2020-unsupervised} model as the encoder, concatenating the source and target sentences as its input for TER/DA score prediction. Both its MonoTransQuest and SiameseTransQuest architectures can achieve good results for sentence-level QE after fine-tuning. Another popular framework, COMET~\citep{rei-etal-2020-comet,stewart-etal-2020-comet} initially relied on reference translation for evaluation, until 2022 when COMETKIWI~\citep{rei-etal-2022-cometkiwi} was introduced to support reference-less evaluation. Similar to MonoTransQuest, it concatenates the source and target, and inputs them into the encoder to get predictions for sentence-level QE scores. 

Given their success in the QE shared tasks in the Conference on Machine Translation (WMT) recently~\citep{specia-etal-2020-findings-wmt,specia-etal-2021-findings,zerva-etal-2022-findings}, TransQuest and COMET are used for fine-tuning to get our baseline models.

The success of LLMs in various natural language processing tasks~\citep{Yang2024} brings new trends and methods in QE research.~\citet{kocmi-federmann-2023-large} proposed a zero-shot prompting technique (called GEMBA) for direct assessment (score from $0$ to $100$) using GPT-4~\citep{OpenAI2023}. They claimed that LLMs without fine-tuning can achieve results comparable to SoTA QE models in score prediction. They further explored the explainability of LLMs in error span detection, and achieved state-of-the-art accuracy for QE system ranking using GPT-4~\citep{kocmi-federmann-2023-gemba}. Based on the GEMBA prompt,~\citet{fernandes-etal-2023-devil} proposed to use LLMs for both score prediction and error categorization. They employed ICL and fine-tuning of LLMs and achieved better results than fine-tuning (encoder-based) multilingual PTLMs. However, fine-tuning LLMs is not cost-effective and energy-efficient. In addition, it might have \textit{catastrophic forgetting}, where a language model \textit{forgets} the knowledge learned during pre-training as it adapts to task-specific data~\citep{McCloskey1989,Ruiz-Garcia2022}. 

Therefore, in this paper, we explore whether PEFT and ICL yield superior performance compared to fine-tuning multilingual PTLMs on the evaluation of machine translation of emotion-loaded UGC. 

\begin{figure*}[h]
\centering
  \begin{minipage}{15cm}
    Score the following translation from Chinese to English with respect to the preservation of emotion on a continuous scale from $0$ to $-100$, \textit{where a score of minus one hundred means ``emotions are critically damaged in multiple places in the text'' and score of zero means ``perfect emotion preservation''. A score of $-1$ means ``very subtle difference in emotion between the source and the target''.} If the score is not zero (not perfect translation), please list keywords or parts of sentences in both source and target where translation is incorrect.
    
    Chinese source: \{Source\_text\} \\
    English translation: \{Machine\_translation\} \\
    The score in terms of emotion preservation for the translation is: \{MQM\_score\} \\
  \end{minipage}
  \caption{Prompt Template 1}
\label{template1}
\end{figure*}

\begin{figure*}[h]
\centering
  \begin{minipage}{15cm}
    Score the following translation from Chinese to English with respect to errors in the preservation of emotion. \textit{The score is calculated based on the number of errors and the level of error severity and weights assigned to each severity level, that is, minor, major and critical. One minor error in emotion preservation, leading to the slight change of emotion after translation, gets a score of $-1$; one major error, pertaining to the change of emotion into a different category after translation, gets a score of $-5$; and one critical error, resulting in the change of emotion into an extremely different or even opposite category after translation, gets a score of $-10$. If there is no error in terms of emotion preservation, the score is 0, which means ``perfect emotion preservation''.} We set a score of $-100$ as the worst score, which means ``there are more than 10 critical errors in emotion preservation''. If the score is not $0$ (imperfect translation), please list keywords or parts of sentences in both source and target where error occurs.
    
    Chinese source: \{Source\_text\} \\
    English translation: \{Machine\_translation\} \\
    The score in terms of emotion preservation for the translation is: \{MQM\_score\} \\
  \end{minipage}
  \caption{Prompt Template 2}
  \label{template2}
\end{figure*}

\section{Data} \label{data}

This section introduces the emotion-related dataset and our extension of QE scores based on MQM. 

\subsection{Emotion-related QE Dataset}

In this paper, we utilize the Human Annotated Dataset for Quality Assessment of Emotion Translation (HADQAET)\footnote{\url{https://github.com/surrey-nlp/HADQAET}} as the main resource~\citep{qian-etal-2023-evaluation}. Its source text originates from the dataset released by the~\textit{Evaluation of Weibo Emotion Classification Technology on the Ninth China National Conference on Social Media Processing} (SMP2020-EWECT) and contains 34,768 instances. Each instance is a tweet-like text segment\footnote{Like most NLP tasks, we treat tweet-like text segments as sentence-level data. However, in contrast to tweets, our instances are longer with an average of $40$ Chinese characters.}, which was manually annotated with one of the six emotion labels, \textit{i.e.}, \textit{anger}, \textit{joy}, \textit{sadness}, \textit{surprise}, \textit{fear} and \textit{neutral}~\citep{Guo2021}. We randomly selected $5,538$ instances with~\textit{non-neutral} emotion labels and used Google Translate for English translation. We proposed an emotion-related MQM framework and recruited two professional translators to annotate errors and their corresponding severity in terms of emotion preservation. Details of our framework, error definition\footnote{The error definition in our prompt templates in Section \ref{in-context} mainly derives from from~\citet{qian-etal-2023-evaluation}.}, error annotation (including inter-annotator agreement), error analysis and data distribution can be seen in~\citet{qian-etal-2023-evaluation}. 

\subsection{Calculation of MQM Scores}

Since \citet{qian-etal-2023-evaluation} only annotated and analyzed the translation errors (and error severity levels) according to the MQM framework, no evaluation score was calculated and proposed. We followed~\citet{freitag-etal-2021-experts} to sum up all weighted errors based on their corresponding severity. The weights for severity levels, as suggested by MQM, are $1$ for \textit{minor error}, $5$ for \textit{major} and $10$ for \textit{critical}. To test the sensitivity of these weights to the overall quality evaluation score, we selected three sets of weights (as shown in Table~\ref{tab:rank-stab}) to check the ranking stability compared with the MQM suggestion. We generated two subsets of $5,000$ instances by sampling with replacement. Then, we calculated the MQM scores using the listed sets of weights. Next, we ranked the scores in ascending order and assessed the similarity of the rankings using the Spearman correlation score~\citep{Spearman1904}. We did this for 1000 times and averaged the ranking similarity. Results are shown in Table~\ref{tab:rank-stab}. 

\begin{table}[H]
\centering
\resizebox{7cm}{!}{%
\begin{tabular}{lc}
\hline
\multicolumn{1}{c}{Sets of Weights} & Ranking Similarities \\ \hline
Minor: $1$, Major: $5$, Critical: $10$ & $0.2711$ \\
Minor: $1$, Major: $3$, Critical: $9$ & $0.0527$ \\
Minor: $1$, Major: $5$, Critical: $15$ & $0.0486$ \\
Minor: $1$, Major: $5$, Critical: $25$ & $0.0515$ \\
\hline
\end{tabular}%
}
\caption{Ranking stability of severity weights}
\label{tab:rank-stab}
\end{table}

From Table~\ref{tab:rank-stab}, we see that the weights suggested by MQM have the highest Spearman correlation score. That means the MQM scores calculated by these weights are most stable. Meanwhile, this set of weights results in a range of scores between $-100$ to $0$, where $-100$ stands for the worst emotion preservation and $0$ for the perfect emotion preservation. The nice range of scores enables us to use prompts designed for DA score prediction such as the GEMBA prompt. 

The calculated MQM scores serve as the true labels for comparison against the predicted scores extracted from the LLM output in both ICL and PEFT scenarios. The source texts and GT translations are utilized to create prompts for the LLM input, as described in Section~\ref{in-context}.

\section{Methodology} \label{methods}

This section explains the methods we used, \textit{i.e.}, ICL and PEFT, with the experimental setup. Selected LLMs and baseline models are listed in Section~\ref{models}. 

\subsection{In-context Learning} \label{in-context}

We devised two prompt templates that include instructions, source text, machine translation and prompt for scores, to ask LLMs to give a score prediction with error explanations. The main difference between our Template 1 (Figure~\ref{template1}) and Template 2 (Figure~\ref{template2}) is the (italic) instruction. Template 1 instructs LLMs to score the machine translation between -100 to 0 and list erroneous words based on emotion preservation. In addition to the basic instruction, Template 2 also includes information about the definition of errors and how the score is calculated based on error severity. 

Apart from zero-shot learning, we employed few-shot learning, where 4 examples\footnote{Due to the input length limit of selected LLMs and the long explanations in the examples, we cannot give more examples than 4.} with different MQM score ranges and errors were inserted into both templates for quality estimation. 

\subsection{PEFT of LLMs}

To maintain model effectiveness while reducing computational costs, we utilized Low-Rank Adaptation (LoRA)~\citep{Hu2021-sv} for parameter efficient fine-tuning of $4$-bit quantized LLMs~\citep{Dettmers2023-dz} instead of full fine-tuning. Both zero-shot and few-shot learning were applied to the fine-tuned LLMs. 

\subsection{Models} \label{models} \label{models}

We selected a wide range of LLMs, mainly open-source models for both ICL and PEFT. Our models include one of the most influential open-source LLMs—Llama-2-13B~\citep{Touvron2023-gw}, models that are claimed to be SoTA Chinese-English LLMs, \textit{i.e.}, Yi-34B\footnote{\url{https://www.01.ai/}} and DeepSeek-67B\footnote{\url{https://www.deepseek.com/}}, and the Mixture-of-Expert (MoE) model, Mixtral-8x7B~\citep{Jiang2024-gx}. Gemini Pro\footnote{\url{https://gemini.google.com/app} at April, 2024}~\citep{Gemini2024} was included in the ICL scenario, to test how proprietary LLMs perform in quality estimation of machine translation of UGC. For PEFT, we tested both the base and the instruction-tuned (chat) models in our experiments. 

\begin{table*}[h]
\centering
\resizebox{10.5cm}{!}{%
\begin{tabular}{cccccc}
\toprule
\multicolumn{2}{c}{Methods} & \multicolumn{2}{c}{Zero-shot Learning} & \multicolumn{2}{c}{Few-shot Learning} \\
Models & Template & $\rho$ & \textit{r} & $\rho$ & \textit{r} \\ \hline
\multirow{2}{*}{Llama-2-13B} & 1 & $0.2143$ & $0.1782$ & $\textit{-0.025}$ & $\textit{-0.0194}$ \\
 & 2 & $\textit{-0.0310}$ & $\textit{0.0260}$ & $\textit{0.0480}$ & $\textit{0.0518}$ \\ \cdashline{2-6}
\multirow{2}{*}{Yi-34B} & 1 & $0.2195$ & $0.1851$ & $0.3470$ & $0.0248$ \\
 & 2 & $0.2060$ & $0.0287$ & $0.3127$ & $0.0236$ \\ \cdashline{2-6}
\multirow{2}{*}{DeepSeek-67B} & 1 & $0.3196$ & $0.1821$ & $\textbf{0.4165}$ & $\textbf{0.2959}$ \\
 & 2 & $0.1956$ & $0.0260$ & $0.3673$ & $0.0294$ \\ \cdashline{2-6}
\multirow{2}{*}{Mixtral-8x7B} & 1 & $0.3154$ & $0.2633$ & $0.3670$ & $0.2870$ \\
 & 2 & $\textbf{0.3484}$ & $\textbf{0.3064}$ & $0.2536$ & $0.0405$ \\ \cdashline{2-6}
\multirow{2}{*}{Gemini Pro} & 1 & $0.2232$ & $0.2416$ & $0.3089$ & $0.1830$ \\
& 2 & $0.2554$ & $0.1833$ & $0.3498$ & $0.2441$ \\
\bottomrule
\end{tabular}%
}
\caption{Spearman $\rho$ and Pearson's~\textit{r} correlation scores for score prediction in \textbf{ICL} scenario}
\label{tab:incontext-learning}
\end{table*}

\begin{table*}[h]
\centering
\resizebox{11cm}{!}{%
\begin{tabular}{cccccc}
\toprule
\multicolumn{2}{c}{Methods} & \multicolumn{2}{c}{Zero-shot Learning} & \multicolumn{2}{c}{Few-shot Learning} \\
Models & Template & \multicolumn{1}{c}{$\rho$} & \multicolumn{1}{c}{\textit{r}} & \multicolumn{1}{c}{$\rho$} & \multicolumn{1}{c}{\textit{r}} \\ \hline
\multirow{2}{*}{Llama-2-13B Chat} & 1 & $0.3114$ & $0.2511$ & $0.1028$ & $0.0061$ \\
 & 2 & $0.3362$ & $0.2782$ & $0.1713$ & $0.1538$ \\ \cdashline{2-6}
 \multirow{2}{*}{Yi-34B Chat} & 1 & $0.5880$ & $0.5902$ & $0.4950$ & $0.3685$ \\
 & 2 & $0.5934$ & $0.5490$ & $\textbf{0.5779}$ & $0.4663$ \\ \cdashline{2-6}
\multirow{2}{*}{DeepSeek-67B Chat} & 1 & $0.5741$ & $0.5325$ & $0.5601$ & $0.5261$ \\
 & 2 & $0.6192$ & $\textbf{0.5983}$ & $0.5567$ & $\textbf{0.5321}$ \\ \cdashline{2-6}
\multirow{2}{*}{Mixtral-8x7B Instruct} & 1 & $0.4577$ & $0.4717$ & $0.4477$ & $0.3444$ \\ 
 & 2 & $0.4256$ & $0.3542$ & $0.3712$ & $0.2709$ \\ \hline
\multirow{2}{*}{Llama-2-13B Base} & 1 & $0.2468$ & $0.3197$ & $0.1371$ & $0.0989$ \\
 & 2 & $0.2848$ & $0.3391$ & $0.0085$ & $0.0226$ \\ \cdashline{2-6}
  \multirow{2}{*}{Yi-34B Base} & 1 & $0.5694$ & $0.4881$ & $0.3589$ & $0.3370$ \\
 & 2 & $0.4883$ & $0.4953$ & $0.2229$ & $0.2286$ \\ \cdashline{2-6}
\multirow{2}{*}{DeepSeek-67B Base} & 1 & $\textbf{0.6498}$ & $0.5433$ & $0.4888$ & $0.4012$ \\
 & 2 & $0.6034$ & $0.5494$ & $0.4350$ & $0.3574$ \\ \cdashline{2-6}
\multirow{2}{*}{Mixtral-8x7B Base} & 1 & $0.4969$ & $0.3125$ & $0.4958$ & $0.4694$ \\
 & 2 & $0.4216$ & $0.3210$ & $0.4530$ & $0.3172$ \\
 \bottomrule
\end{tabular}%
}
\caption{Spearman $\rho$ and Pearson's~\textit{r} correlation scores for score prediction in \textbf{PEFT} scenario}
\label{tab:peft-score}
\end{table*}

\paragraph{Baselines}

We utilized TransQuest (including MonoTransQuest and SiameseTransQuest) and COMET to fine-tune multilingual PTLMs like XLM-RoBERTa as our baselines. We also continued fine-tuning on HADQAET after we fine-tuned XLM-RoBERTa$_{large}$ on the Chinese-English sentence-level MQM dataset from WMT20-22~\citep{freitag-etal-2021-experts,freitag-etal-2021-results,freitag-etal-2022-results}. 

\subsection{Experimental Setup}

We evaluated the two prompt templates on the models listed in Section~\ref{models}, focusing on score prediction with error explanations. The evaluation was conducted under both ICL and PEFT scenarios, using zero-shot and few-shot learning approaches. The predicted scores were extracted from the LLM-generated texts using regular expression. They were evaluated using Spearman $\rho$ and Pearson's ~\textit{r} correlation scores. 

We divided the data into training, validation, and test sets in proportions of $80\%$, $10\%$, and $10\%$. Baseline models were fine-tuned for $2$ epochs with a learning rate of $2e-5$, batch size of $8$ and sequence length of $200$ on an NVIDIA Quadro RTX 5000 GPU. For LLM inference, the temperature hyperparameter was set as $0.95$ and top\_p as $0.7$. All LLMs were loaded in 4-bits using LLaMA-Factory~\citep{zheng2024llamafactory} for both inference and PEFT. For PEFT, we chose the rank to be $8$, alpha to be $64$, and the target layers to be the attention layers based on experimentation. All LLMs were trained for $3$ epochs with a learning rate of $5e-5$ and a batch size of $4$ using an NVIDIA A40 GPU. 

\begin{table}[h]
\centering
\resizebox{7cm}{!}{%
\begin{tabular}{ccc}
\toprule
Methods & $\rho$ & \textit{r} \\ \hline
MonoTransQuest (FT) & $0.4355$ & $0.3984$ \\
SiameseTransQuest (FT) & $0.4151$ & $0.4502$ \\
COMET (FT) & $0.4083$ & $0.3699$ \\
MonoTransQuest (CFT) & $0.4527$ & $0.4050$ \\
SiameseTransQuest (CFT) & $\textbf{0.5118}$ & $\textbf{0.4934}$ \\
\bottomrule
\end{tabular}%
}
\caption{Spearman $\rho$ and Pearson's~\textit{r} correlation scores of baseline models fine-tuned (FT) and continued fine-tuned (CFT) using TransQuest and COMET. CFT models are obtained by fine-tuning on WMT MQM data first and then continued fine-tuning on HADQAET and results are evaluated on HADQAET.}
\label{tab:ft}
\end{table}

\begin{CJK*}{UTF8}{gbsn}
\begin{figure*}[h]
\centering
  \begin{minipage}{15cm}
    Source: 我下次洗衣服的时候如果再放很多洗衣液进去我就是狗，怎么洗都还是很多泡沫。FUCK! \\
    GT output: If I put a lot of laundry detergent in the next time I do laundry, I will be a dog, and I will still have a lot of foam no matter how I wash it. FUCK! \\
    Human translation: If I still use a lot of laundry detergent in the next time I do laundry, I will be a dog. No matter how I wash it, there is still lots of foam. FUCK! 
  \end{minipage}
  \caption{An example of refusal to reply because of interjections}
  \label{failure1}
\end{figure*}
\end{CJK*}

\begin{CJK*}{UTF8}{gbsn}
\begin{figure*}[h]
\centering
  \begin{minipage}{15cm}
    Source: 踢个香港，裁判黑成这样…不知道该说什么了 \\
    GT output: Kick Hong Kong, the referee is so black... I don't know what to say \\
    Human translation: In the (football) game against Hong Kong, the referee was so biased!!! Don't know what to say. \\
    LLM output: I apologize, but I cannot provide a score for the translation as it is not appropriate to use language that is derogatory or offensive. The phrase~\textbf{``裁判黑'' (referee is black) is not a neutral or factual statement, and it could be perceived as racist or discriminatory}. Additionally, the phrase~\textbf{``踢个香港'' (kick Hong Kong) is also not appropriate, as it could be interpreted as advocating for violence or aggression towards a specific group of people}. Instead, I would suggest using more neutral and respectful language to convey the intended meaning. For example, the phrase ``裁判是黑的'' (the referee is black) could be used to describe the physical appearance of the referee, without any negative or derogatory connotations. Similarly, the phrase ``踢个球'' (kick the ball) could be used to describe the action of kicking a ball, without any reference to a specific location or group of people. It's important to be mindful of the language we use and the impact it can have on others. Using respectful and inclusive language is essential for maintaining a positive and welcoming environment for everyone.
  \end{minipage}
  \caption{An example of refusal to reply because of ``sensitive'' words}
  \label{failure2}
\end{figure*}
\end{CJK*}

\section{Results and Manual Analysis} \label{results}

This section shows the results under our ICL and PEFT scenarios in Tables \ref{tab:incontext-learning} and \ref{tab:peft-score} in comparison with the baselines in Table \ref{tab:ft}. A manual analysis of some LLM outputs is presented in $\S$~\ref{manual_ana}. 

\subsection{Baselines}

Table~\ref{tab:ft} shows results of simple and continued fine-tuning on HADQAET and the Chinese-English MQM dataset from WMT20-22 using TransQuest and COMET. We observe that TransQuest achieved higher correlation scores than COMET when fine-tuned on HADQAET. Therefore, only TransQuest was used for continued fine-tuning. The highest Spearman and Pearson correlation scores $0.5118$ and $0.4934$ were achieved by continued fine-tuning of models that had been fine-tuned on the MQM dataset using SiameseTransQuest. 

\subsection{In-context Learning}

Table~\ref{tab:incontext-learning} shows ICL results using the two prompt templates. We cannot see a clear performance boost using Template 2 where error severities are clearly defined and described. In contrast, we see an obvious drop in correlation scores using Template 2 for Llama-2-13B (see Section~\ref{manual_ana} for our investigation). Few-shot learning improves model performance\footnote{Few-shot results are bad for Llama-2-13B, because it cuts off the prompt while reaching the input length limit.} like the cases for Yi-34B, DeepSeek-67B and Gemini Pro.

The MoE model, Mixtral-8x7B achieves the highest correlation scores in zero-shot setting, while DeepSeek-67B is the best at few-shot learning, obtaining a Spearman correlation score of 0.4165, very close to the fine-tuned models using TransQuest and COMET. Gemini Pro, the proprietary LLM, failed to achieve higher correlation scores than the open-source ones in this task. ICL results of all above models are no better than our continued fine-tuning baselines. 

\begin{CJK*}{UTF8}{gbsn}
\begin{figure*}[h]
\centering
  \begin{minipage}{15cm}
    \textbf{Case 1} \\
    I would score this translation as follows: \\
    Minor errors: 1; Major errors: 1; Critical errors: 0; Total score: -2 \\
    Here's a breakdown of the errors and their severity levels: \\
    1. Minor error: The word ``满满'' (full of) in the source text is not accurately conveyed in the target text as ``full of love''. A more accurate translation would be ``very full'' or ``overwhelmingly full''. (Score: -1) \\
    2. Major error: The word ``爱'' (love) in the target text is not present in the source text. This is a significant deviation from the original meaning, as the source text does not mention love. (Score: -5) Therefore, the total score for this translation is -2. \\
    In terms of error locations, the minor error occurs in the word ``满满'' not being accurately conveyed, and the major error occurs in the word ``爱'' being added to the target text despite not being present in the source text. \\
    \textbf{Case 2} \\
    I'd be happy to help! Here's the translation of the Chinese sentence into English, with an error analysis based on the criteria you provided: \\
    Chinese source: 遇到你和你在一起是我最大最大的荣幸 \\
    English translation: It is my greatest honor to meet you and be with you \\
    Error analysis: Minor error in emotion preservation: The word ``荣幸'' (rangxing) in the Chinese source is not perfectly preserved in the English translation as ``honor''. While ``honor'' does convey a sense of pride and respect, it does not fully capture the nuance of ``荣幸'', which carries a stronger connotation of joy and happiness. Therefore, the score for this translation is -1. \\
    Keywords or parts of sentences where error occurs: Chinese: 遇到你和你在一起 English: It is my greatest honor to meet you and be with you \\
    Note: The word ``和'' (he) in the Chinese source is not translated in the English translation, which could potentially result in a loss of information and context. However, this omission does not significantly affect the overall meaning of the sentence, so it is not counted as an error in this case.
  \end{minipage}
  \caption{An example of unstable output}
  \label{failure3}
\end{figure*}
\end{CJK*}

\begin{figure*}[h]
\centering
  \begin{minipage}{15cm}
    \textbf{Output 1} \\
    I'd be happy to help you with that! ...... The score is based on the severity level of the errors and the number of errors. In this case, there is only one minor error, so the score is $-1$.

    \textbf{Output 2} \\
    I'd be happy to help you with that! ...... The score is based on the severity of the errors and the level of emotion preservation. A score of $-10$ indicates a critical error in emotion preservation, while a score of $-1$ indicates a minor error. A score of $0$ indicates perfect emotion preservation. 
  \end{minipage}
  \caption{Different outputs using the exact same prompt and hyperparameters (temperature as $0$) from Llama-2-13B, where the same repetitive text in both outputs is omitted and denoted as ``......''.}
  \label{failure4}
\end{figure*}

\subsection{PEFT}

Table~\ref{tab:peft-score} presents results for score prediction in the PEFT scenario. We observe that while few-shot learning usually results in better performance in ICL scenario, its performance is worse than zero-shot post-PEFT, especially for base models. We observe that the performance drop for base models after fine-tuning is more obvious than instruction-tuned models in the few-shot setting, with the exception of Mixtral-8x7B. The findings in ICL indicate that the MoE model outperforms regular dense models of similar size. We anticipated that the Mixtral-8x7B model would yield significantly improved results after PEFT, but the observed enhancement was not as substantial as expected. We attained our highest correlation scores of $0.6498$ and $0.5983$ by fine-tuning the DeepSeek models, with both Spearman and Pearson correlation scores surpassing the baselines. These results underscore the effectiveness of PEFT for LLMs towards achieving state-of-the-art performance in quality estimation. 

\subsection{Manual Analysis} \label{manual_ana}

While most Spearman correlation scores are positive and larger than $0.1$, it is noteworthy that Llama-2-13B outputs QE scores that exhibit a negative correlation ($-0.0310$) with the true scores using Template 2. For further investigation, we did a manual analysis of the model output with the help of a Chinese-English translator. We observe two phenomena that might pose challenges for using LLMs to evaluate translation quality: \textit{1) refusal to reply} because of ``inappropriate language'', and \textit{2) unstable output} patterns.

\subsubsection{Refusal to Reply}

We find Llama-2-13B \textit{refused to evaluate} $4.97\%$ of the instances\footnote{They were excluded for correlation score computation.} in the test set, because the source texts contain swear words from social media. However, most of these words are used as interjections to express the \textit{angry} emotion of the blogger towards a certain event as shown in Figure~\ref{failure1}, not aggression towards someone. Llama-2-13B seems to refuse to answer any questions containing these words. 

\begin{CJK*}{UTF8}{gbsn}
Of particular interest, Llama-2-13B demonstrates heightened sensitivity to language associated with discrimination and aggression. As shown in Figure~\ref{failure2}, the Chinese source text complains about a football game against Hong Kong. It mentions ``踢''~\textit{kick (ball)} and ``香港''~\textit{Hong Kong}, which Llama-2-13B believes it could be interpreted as ``advocating for violence or aggression towards a specific group of people''. ``裁判黑'' in the source means the referee manipulates the game, as the character ``黑'', which has the meaning of ``black'', means doing something behind the scenes in this context. Llama-2-13B is over-sensitive about using the character ``黑'' to describe a person. This may become a problem for evaluating translation quality, especially emotion-load UGC. 
\end{CJK*}

\subsubsection{Unstable Output}

We expect LLMs to output texts with similar structures or patterns when the same prompt template is used. However, responses from Llama-2-13B sometimes varied. Some answers indicate a misunderstanding of the instruction in the prompt, whereas others seem to follow the instruction and perform the quality evaluation task. 

As shown in Figure~\ref{failure3}, the output structure of Case 1 and Case 2 are very different even when using the same prompt template. In Case 1, Llama-2-13B lists the number of errors based on severity levels and generates a total score, which is inconsistent with its following analysis. The analysis thereafter breaks down the errors and provides a score to each error, but the total score is calculated \textit{incorrectly} due to its poor reasoning ability ~\citep{Arkoudas2023}. In Case 2, Llama-2-13B starts with error analysis and then produces a total score without mentioning scores for each error. 

We observe unstable output even when the temperature hyperparameter is set to zero, which essentially eliminates the sampling process and is supposed to produce the exact same consistent output. However, as shown in Figure~\ref{failure4}, we observe variance in outputs from Llama-2-13B after prompting with the same text several times, using identical hyperparameters ($0$ temperature). Inconsistent output structures might cause problems for extracting the QE scores for the computation of the overall correlation, and more importantly, confuse users in understanding the real translation quality. 

The phenomena of refusal to reply and unstable output were not observed only in the Llama-2-13B model. Other LLMs might also refuse to reply to questions containing swear words and output inconsistent text structures. Interestingly, we find that models proposed by Chinese companies such as Yi and DeepSeek are less sensitive to words related to discrimination and aggression, unlike Llama and ChatGPT, as they usually provide a QE score to such examples. However, this needs to be verified by further experiments using more LLMs.

\section{Conclusion} \label{conclusion}

In order to understand whether LLMs are state-of-the-art quality estimators for machine translation of emotion-loaded UGC, our paper utilized an existing emotion-related dataset with human-annotated errors. We computed the MQM scores based on the translation errors, and devised two prompt templates to allow LLMs to perform score prediction with error explanations. Different types and sizes of LLMs were employed to compare with fine-tuning of multilingual PTLMs, under ICL and PEFT scenarios. We find that while LLMs can obtain good correlation scores in zero-shot setting, PEFT of LLMs leads to state-of-the-art performance in score prediction with error explanations, which resolves the \textit{opacity issue} of current QE models. However, a manual analysis reveals that LLMs still have problems such as refusal to reply and unstable output while performing the QE task. Users need to be mindful when using LLMs for quality evaluation. For future work, we will investigate how LLMs perform on the evaluation of general MT quality under ICL and PEFT scenarios. 

\section{Limitations} \label{limits}

Our experimentation is limited to a small number of LLMs listed in Section~\ref{models}, due to the economic, time and energy cost in LLM training and inferencing. Results might be different on other LLMs.  Meanwhile, although LLM-based evaluation is more interpretable and accurate, it is much more time- and energy-consuming than using regular QE models. 

\bibliography{custom}

\begin{thebibliography}{38}
\providecommand{\natexlab}[1]{#1}

\bibitem[{Arkoudas(2023)}]{Arkoudas2023}
Konstantine Arkoudas. 2023.
\newblock \href {https://arxiv.org/abs/2308.03762} {{GPT-4} can't reason}.
\newblock \emph{arXiv preprint}, arXiv:2308.03762.

\bibitem[{Conneau et~al.(2020)Conneau, Khandelwal, Goyal, Chaudhary, Wenzek, Guzm{\'a}n, Grave, Ott, Zettlemoyer, and Stoyanov}]{conneau-etal-2020-unsupervised}
Alexis Conneau, Kartikay Khandelwal, Naman Goyal, Vishrav Chaudhary, Guillaume Wenzek, Francisco Guzm{\'a}n, Edouard Grave, Myle Ott, Luke Zettlemoyer, and Veselin Stoyanov. 2020.
\newblock \href {https://doi.org/10.18653/v1/2020.acl-main.747} {Unsupervised cross-lingual representation learning at scale}.
\newblock In \emph{Proceedings of the 58th Annual Meeting of the Association for Computational Linguistics}, pages 8440--8451, Online. Association for Computational Linguistics.

\bibitem[{Dettmers et~al.(2023)Dettmers, Pagnoni, Holtzman, and Zettlemoyer}]{Dettmers2023-dz}
Tim Dettmers, Artidoro Pagnoni, Ari Holtzman, and Luke Zettlemoyer. 2023.
\newblock \href {https://proceedings.neurips.cc/paper_files/paper/2023/file/1feb87871436031bdc0f2beaa62a049b-Paper-Conference.pdf} {{QLoRA: Efficient Finetuning of Quantized LLMs}}.
\newblock In \emph{Advances in Neural Information Processing Systems}, volume~36, pages 10088--10115. Curran Associates, Inc.

\bibitem[{Dorr et~al.(2011)Dorr, Olive, McCary, and Christianson}]{Dorr2011}
Bonnie Dorr, Joseph Olive, John McCary, and Caitlin Christianson. 2011.
\newblock \href {https://doi.org/10.1007/978-1-4419-7713-7\_5} {{Machine Translation Evaluation and Optimization}}.
\newblock In J.~Olive, C.~Christianson, and J.~McCary, editors, \emph{Handbook of Natural Language Processing and Machine Translation}, pages 745--843. Springer.

\bibitem[{Fernandes et~al.(2023)Fernandes, Deutsch, Finkelstein, Riley, Martins, Neubig, Garg, Clark, Freitag, and Firat}]{fernandes-etal-2023-devil}
Patrick Fernandes, Daniel Deutsch, Mara Finkelstein, Parker Riley, Andr{\'e} Martins, Graham Neubig, Ankush Garg, Jonathan Clark, Markus Freitag, and Orhan Firat. 2023.
\newblock \href {https://doi.org/10.18653/v1/2023.wmt-1.100} {The devil is in the errors: Leveraging large language models for fine-grained machine translation evaluation}.
\newblock In \emph{Proceedings of the Eighth Conference on Machine Translation}, pages 1066--1083, Singapore. Association for Computational Linguistics.

\bibitem[{Freitag et~al.(2021{\natexlab{a}})Freitag, Foster, Grangier, Ratnakar, Tan, and Macherey}]{freitag-etal-2021-experts}
Markus Freitag, George Foster, David Grangier, Viresh Ratnakar, Qijun Tan, and Wolfgang Macherey. 2021{\natexlab{a}}.
\newblock \href {https://doi.org/10.1162/tacl\_a\_00437} {Experts, errors, and context: A large-scale study of human evaluation for machine translation}.
\newblock In \emph{Transactions of the Association for Computational Linguistics}, volume~9, pages 1460--1474, Cambridge, MA. MIT Press.

\bibitem[{Freitag et~al.(2022)Freitag, Rei, Mathur, Lo, Stewart, Avramidis, Kocmi, Foster, Lavie, and Martins}]{freitag-etal-2022-results}
Markus Freitag, Ricardo Rei, Nitika Mathur, Chi-kiu Lo, Craig Stewart, Eleftherios Avramidis, Tom Kocmi, George Foster, Alon Lavie, and Andr{\'e} F.~T. Martins. 2022.
\newblock \href {https://aclanthology.org/2022.wmt-1.2} {Results of {WMT}22 metrics shared task: Stop using {BLEU} {--} neural metrics are better and more robust}.
\newblock In \emph{Proceedings of the Seventh Conference on Machine Translation (WMT)}, pages 46--68, Abu Dhabi, United Arab Emirates (Hybrid). Association for Computational Linguistics.

\bibitem[{Freitag et~al.(2021{\natexlab{b}})Freitag, Rei, Mathur, Lo, Stewart, Foster, Lavie, and Bojar}]{freitag-etal-2021-results}
Markus Freitag, Ricardo Rei, Nitika Mathur, Chi-kiu Lo, Craig Stewart, George Foster, Alon Lavie, and Ond{\v{r}}ej Bojar. 2021{\natexlab{b}}.
\newblock \href {https://aclanthology.org/2021.wmt-1.73} {Results of the {WMT}21 metrics shared task: Evaluating metrics with expert-based human evaluations on {TED} and news domain}.
\newblock In \emph{Proceedings of the Sixth Conference on Machine Translation}, pages 733--774, Online. Association for Computational Linguistics.

\bibitem[{{Gemini Team}(2024)}]{Gemini2024}
{Gemini Team}. 2024.
\newblock \href {https://arxiv.org/abs/2403.05530} {Gemini 1.5: Unlocking multimodal understanding across millions of tokens of context}.
\newblock \emph{arXiv preprint}, arXiv:2403.05530.

\bibitem[{Graham et~al.(2013)Graham, Baldwin, Moffat, and Zobel}]{graham-etal-2013-continuous}
Yvette Graham, Timothy Baldwin, Alistair Moffat, and Justin Zobel. 2013.
\newblock \href {https://aclanthology.org/W13-2305} {Continuous measurement scales in human evaluation of machine translation}.
\newblock In \emph{Proceedings of the 7th Linguistic Annotation Workshop and Interoperability with Discourse}, pages 33--41, Sofia, Bulgaria. Association for Computational Linguistics.

\bibitem[{Guerreiro et~al.(2024)Guerreiro, Rei, Stigt, Coheur, Colombo, and Martins}]{Guerreiro2023-pf}
Nuno~M. Guerreiro, Ricardo Rei, Daan~van Stigt, Luisa Coheur, Pierre Colombo, and André F.~T. Martins. 2024.
\newblock \href {https://doi.org/10.1162/tacl_a_00683} {{xcomet: Transparent Machine Translation Evaluation through Fine-grained Error Detection}}.
\newblock \emph{Transactions of the Association for Computational Linguistics}, 12:979--995.

\bibitem[{Guo et~al.(2021)Guo, Lai, Xiang, Yu, and Huang}]{Guo2021}
Xianwei Guo, Hua Lai, Yan Xiang, Zhengtao Yu, and Yuxin Huang. 2021.
\newblock \href {https://aclanthology.org/2021.ccl-1.82} {{Emotion Classification of COVID-19 Chinese Microblogs based on the Emotion Category Description}}.
\newblock pages 916--927. Chinese Information Processing Society of China.

\bibitem[{Hassan et~al.(2018)Hassan, Aue, Chen, Chowdhary, Clark, Federmann, Huang, Junczys-Dowmunt, Lewis, Li, Liu, Liu, Luo, Menezes, Qin, Seide, Tan, Tian, Wu, Wu, Xia, Zhang, Zhang, and Zhou}]{Hassan2018}
Hany Hassan, Anthony Aue, Chang Chen, Vishal Chowdhary, Jonathan Clark, Christian Federmann, Xuedong Huang, Marcin Junczys-Dowmunt, William Lewis, Mu~Li, Shujie Liu, Tie-Yan Liu, Renqian Luo, Arul Menezes, Tao Qin, Frank Seide, Xu~Tan, Fei Tian, Lijun Wu, Shuangzhi Wu, Yingce Xia, Dongdong Zhang, Zhirui Zhang, and Ming Zhou. 2018.
\newblock \href {https://arxiv.org/abs/arXiv:1803.05567} {{Achieving Human Parity on Automatic Chinese to English News Translation}}.
\newblock \emph{arXive preprint}.

\bibitem[{Hu et~al.(2022)Hu, yelong shen, Wallis, Allen-Zhu, Li, Wang, Wang, and Chen}]{Hu2021-sv}
Edward~J Hu, yelong shen, Phillip Wallis, Zeyuan Allen-Zhu, Yuanzhi Li, Shean Wang, Lu~Wang, and Weizhu Chen. 2022.
\newblock \href {https://openreview.net/forum?id=nZeVKeeFYf9} {{LoRA: Low-Rank Adaptation of Large Language Models}}.
\newblock In \emph{International Conference on Learning Representations}.

\bibitem[{Jiang et~al.(2024)Jiang, Sablayrolles, Roux, Mensch, Savary, Bamford, Chaplot, Casas, Hanna, Bressand, Lengyel, Bour, Lample, Lavaud, Saulnier, Lachaux, Stock, Subramanian, Yang, Antoniak, Scao, Gervet, Lavril, Wang, Lacroix, and Sayed}]{Jiang2024-gx}
Albert~Q Jiang, Alexandre Sablayrolles, Antoine Roux, Arthur Mensch, Blanche Savary, Chris Bamford, Devendra~Singh Chaplot, Diego de~las Casas, Emma~Bou Hanna, Florian Bressand, Gianna Lengyel, Guillaume Bour, Guillaume Lample, L{\'e}lio~Renard Lavaud, Lucile Saulnier, Marie-Anne Lachaux, Pierre Stock, Sandeep Subramanian, Sophia Yang, Szymon Antoniak, Teven~Le Scao, Th{\'e}ophile Gervet, Thibaut Lavril, Thomas Wang, Timoth{\'e}e Lacroix, and William~El Sayed. 2024.
\newblock \href {https://arxiv.org/abs/2401.04088} {{Mixtral of Experts}}.
\newblock \emph{Preprint}, arXiv:2401.04088.

\bibitem[{Kocmi and Federmann(2023{\natexlab{a}})}]{kocmi-federmann-2023-gemba}
Tom Kocmi and Christian Federmann. 2023{\natexlab{a}}.
\newblock \href {https://doi.org/10.18653/v1/2023.wmt-1.64} {{GEMBA}-{MQM}: Detecting translation quality error spans with {GPT}-4}.
\newblock In \emph{Proceedings of the Eighth Conference on Machine Translation}, pages 768--775, Singapore. Association for Computational Linguistics.

\bibitem[{Kocmi and Federmann(2023{\natexlab{b}})}]{kocmi-federmann-2023-large}
Tom Kocmi and Christian Federmann. 2023{\natexlab{b}}.
\newblock \href {https://aclanthology.org/2023.eamt-1.19} {Large language models are state-of-the-art evaluators of translation quality}.
\newblock In \emph{Proceedings of the 24th Annual Conference of the European Association for Machine Translation}, pages 193--203, Tampere, Finland. European Association for Machine Translation.

\bibitem[{Lai et~al.(2020)Lai, Dai, and Yang}]{Lai2020}
Guokun Lai, Zihang Dai, and Yiming Yang. 2020.
\newblock \href {https://arxiv.org/abs/arXiv:2009.08595} {{Unsupervised Parallel Corpus Mining on Web Data}}.

\bibitem[{Lommel et~al.(2014)Lommel, Burchardt, and Uszkoreit}]{Lommel2014}
Arle~Richard Lommel, Aljoscha Burchardt, and Hans Uszkoreit. 2014.
\newblock \href {https://doi.org/10.5565/rev/tradumatica.77} {{Multidimensional Quality Metrics: A Flexible System for Assessing Translation Quality}}.
\newblock \emph{Tradumàtica: tecnologies de la traducció}, 0:455--463.

\bibitem[{McCloskey and Cohen(1989)}]{McCloskey1989}
Michael McCloskey and Neal~J. Cohen. 1989.
\newblock \href {https://doi.org/10.1016/S0079-7421(08)60536-8} {{Catastrophic Interference in Connectionist Networks: The Sequential Learning Problem}}.
\newblock volume~24, pages 109--165. Academic Press.

\bibitem[{OpenAI(2023)}]{OpenAI2023}
OpenAI. 2023.
\newblock \href {https://arxiv.org/abs/arXiv:2303.08774} {{GPT-4 Technical Report}}.
\newblock \emph{arXive preprint}.

\bibitem[{Qian et~al.(2023)Qian, Orasan, Carmo, Li, and Kanojia}]{qian-etal-2023-evaluation}
Shenbin Qian, Constantin Orasan, Felix~Do Carmo, Qiuliang Li, and Diptesh Kanojia. 2023.
\newblock \href {https://aclanthology.org/2023.eamt-1.13} {Evaluation of {C}hinese-{E}nglish machine translation of emotion-loaded microblog texts: A human annotated dataset for the quality assessment of emotion translation}.
\newblock In \emph{Proceedings of the 24th Annual Conference of the European Association for Machine Translation}, pages 125--135, Tampere, Finland. European Association for Machine Translation.

\bibitem[{Ranasinghe et~al.(2020)Ranasinghe, Orasan, and Mitkov}]{Ranasinghe2020}
Tharindu Ranasinghe, Constantin Orasan, and Ruslan Mitkov. 2020.
\newblock \href {https://aclanthology.org/2020.coling-main.445.pdf} {{TransQuest: Translation Quality Estimation with Cross-lingual Transformers}}.
\newblock pages 5070--5081. International Committee on Computational Linguistics.

\bibitem[{Rei et~al.(2020)Rei, Stewart, Farinha, and Lavie}]{rei-etal-2020-comet}
Ricardo Rei, Craig Stewart, Ana~C Farinha, and Alon Lavie. 2020.
\newblock \href {https://doi.org/10.18653/v1/2020.emnlp-main.213} {{COMET}: A neural framework for {MT} evaluation}.
\newblock In \emph{Proceedings of the 2020 Conference on Empirical Methods in Natural Language Processing (EMNLP)}, pages 2685--2702, Online. Association for Computational Linguistics.

\bibitem[{Rei et~al.(2022)Rei, Treviso, Guerreiro, Zerva, Farinha, Maroti, C.~de Souza, Glushkova, Alves, Coheur, Lavie, and Martins}]{rei-etal-2022-cometkiwi}
Ricardo Rei, Marcos Treviso, Nuno~M. Guerreiro, Chrysoula Zerva, Ana~C Farinha, Christine Maroti, Jos{\'e}~G. C.~de Souza, Taisiya Glushkova, Duarte Alves, Luisa Coheur, Alon Lavie, and Andr{\'e} F.~T. Martins. 2022.
\newblock \href {https://aclanthology.org/2022.wmt-1.60} {{C}omet{K}iwi: {IST}-unbabel 2022 submission for the quality estimation shared task}.
\newblock In \emph{Proceedings of the Seventh Conference on Machine Translation (WMT)}, pages 634--645, Abu Dhabi, United Arab Emirates (Hybrid). Association for Computational Linguistics.

\bibitem[{Ruiz-Garcia(2022)}]{Ruiz-Garcia2022}
Miguel Ruiz-Garcia. 2022.
\newblock \href {https://doi.org/10.1038/s41598-022-14348-x} {{Model architecture can transform catastrophic forgetting into positive transfer}}.
\newblock \emph{Scientific Reports}, 12.

\bibitem[{Saadany et~al.(2023)Saadany, Orasan, Quintana, Carmo, and Zilio}]{saadany-etal-2023-analysing}
Hadeel Saadany, Constantin Orasan, Rocio~Caro Quintana, Felix~Do Carmo, and Leonardo Zilio. 2023.
\newblock \href {https://aclanthology.org/2023.eamt-1.27} {Analysing mistranslation of emotions in multilingual tweets by online {MT} tools}.
\newblock In \emph{Proceedings of the 24th Annual Conference of the European Association for Machine Translation}, pages 275--284, Tampere, Finland. European Association for Machine Translation.

\bibitem[{Snover et~al.(2006)Snover, Dorr, Schwartz, Micciulla, and Makhoul}]{snover-etal-2006-study}
Matthew Snover, Bonnie Dorr, Rich Schwartz, Linnea Micciulla, and John Makhoul. 2006.
\newblock \href {https://aclanthology.org/2006.amta-papers.25} {A study of translation edit rate with targeted human annotation}.
\newblock In \emph{Proceedings of the 7th Conference of the Association for Machine Translation in the Americas: Technical Papers}, pages 223--231, Cambridge, Massachusetts, USA. Association for Machine Translation in the Americas.

\bibitem[{Spearman(1904)}]{Spearman1904}
Charles Spearman. 1904.
\newblock \href {http://www.jstor.org/stable/1412159?origin=JSTOR-pdf} {The proof and measurement of association between two things}.
\newblock \emph{The American Journal of Psychology}, 15:72--101.

\bibitem[{Specia et~al.(2020)Specia, Blain, Fomicheva, Fonseca, Chaudhary, Guzm{\'a}n, and Martins}]{specia-etal-2020-findings-wmt}
Lucia Specia, Fr{\'e}d{\'e}ric Blain, Marina Fomicheva, Erick Fonseca, Vishrav Chaudhary, Francisco Guzm{\'a}n, and Andr{\'e} F.~T. Martins. 2020.
\newblock \href {https://aclanthology.org/2020.wmt-1.79} {Findings of the {WMT} 2020 shared task on quality estimation}.
\newblock In \emph{Proceedings of the Fifth Conference on Machine Translation}, pages 743--764, Online. Association for Computational Linguistics.

\bibitem[{Specia et~al.(2021)Specia, Blain, Fomicheva, Zerva, Li, Chaudhary, and Martins}]{specia-etal-2021-findings}
Lucia Specia, Fr{\'e}d{\'e}ric Blain, Marina Fomicheva, Chrysoula Zerva, Zhenhao Li, Vishrav Chaudhary, and Andr{\'e} F.~T. Martins. 2021.
\newblock \href {https://aclanthology.org/2021.wmt-1.71} {Findings of the {WMT} 2021 shared task on quality estimation}.
\newblock In \emph{Proceedings of the Sixth Conference on Machine Translation}, pages 684--725, Online. Association for Computational Linguistics.

\bibitem[{Specia et~al.(2018)Specia, Scarton, and Paetzold}]{Specia2018}
Lucia Specia, Carolina Scarton, and Gustavo~Henrique Paetzold. 2018.
\newblock \href {https://doi.org/10.1007/978-3-031-02168-8_1} {\emph{Quality Estimation for Machine Translation}}.
\newblock Spinger, Cham, Germany.

\bibitem[{Stewart et~al.(2020)Stewart, Rei, Farinha, and Lavie}]{stewart-etal-2020-comet}
Craig Stewart, Ricardo Rei, Catarina Farinha, and Alon Lavie. 2020.
\newblock \href {https://aclanthology.org/2020.amta-user.4} {{COMET} - deploying a new state-of-the-art {MT} evaluation metric in production}.
\newblock In \emph{Proceedings of the 14th Conference of the Association for Machine Translation in the Americas (Volume 2: User Track)}, pages 78--109, Virtual. Association for Machine Translation in the Americas.

\bibitem[{Touvron et~al.(2023)Touvron, Martin, Stone, Albert, Almahairi, Babaei, Bashlykov, Batra, Bhargava, Bhosale, Bikel, Blecher, Ferrer, Chen, Cucurull, Esiobu, Fernandes, Fu, Fu, Fuller, Gao, Goswami, Goyal, Hartshorn, Hosseini, Hou, Inan, Kardas, Kerkez, Khabsa, Kloumann, Korenev, Koura, Lachaux, Lavril, Lee, Liskovich, Lu, Mao, Martinet, Mihaylov, Mishra, Molybog, Nie, Poulton, Reizenstein, Rungta, Saladi, Schelten, Silva, Smith, Subramanian, Tan, Tang, Taylor, Williams, Kuan, Xu, Yan, Zarov, Zhang, Fan, Kambadur, Narang, Rodriguez, Stojnic, Edunov, and Scialom}]{Touvron2023-gw}
Hugo Touvron, Louis Martin, Kevin Stone, Peter Albert, Amjad Almahairi, Yasmine Babaei, Nikolay Bashlykov, Soumya Batra, Prajjwal Bhargava, Shruti Bhosale, Dan Bikel, Lukas Blecher, Cristian~Canton Ferrer, Moya Chen, Guillem Cucurull, David Esiobu, Jude Fernandes, Jeremy Fu, Wenyin Fu, Brian Fuller, Cynthia Gao, Vedanuj Goswami, Naman Goyal, Anthony Hartshorn, Saghar Hosseini, Rui Hou, Hakan Inan, Marcin Kardas, Viktor Kerkez, Madian Khabsa, Isabel Kloumann, Artem Korenev, Punit~Singh Koura, Marie-Anne Lachaux, Thibaut Lavril, Jenya Lee, Diana Liskovich, Yinghai Lu, Yuning Mao, Xavier Martinet, Todor Mihaylov, Pushkar Mishra, Igor Molybog, Yixin Nie, Andrew Poulton, Jeremy Reizenstein, Rashi Rungta, Kalyan Saladi, Alan Schelten, Ruan Silva, Eric~Michael Smith, Ranjan Subramanian, Xiaoqing~Ellen Tan, Binh Tang, Ross Taylor, Adina Williams, Jian~Xiang Kuan, Puxin Xu, Zheng Yan, Iliyan Zarov, Yuchen Zhang, Angela Fan, Melanie Kambadur, Sharan Narang, Aurelien Rodriguez, Robert Stojnic, Sergey Edunov, and Thomas
  Scialom. 2023.
\newblock \href {https://arxiv.org/abs/2307.09288} {{Llama 2: Open foundation and fine-tuned chat models}}.
\newblock \emph{Preprint}, arXiv:2307.09288.

\bibitem[{Wang et~al.(2021)Wang, Tu, Tan, Wang, Sun, and Liu}]{Wang2021-lh}
Shuo Wang, Zhaopeng Tu, Zhixing Tan, Wenxuan Wang, Maosong Sun, and Yang Liu. 2021.
\newblock \href {https://arxiv.org/abs/2106.13627} {Language models are good translators}.
\newblock \emph{arXiv preprint}.

\bibitem[{Yang et~al.(2024)Yang, Jin, Tang, Han, Feng, Jiang, Zhong, Yin, and Hu}]{Yang2024}
Jingfeng Yang, Hongye Jin, Ruixiang Tang, Xiaotian Han, Qizhang Feng, Haoming Jiang, Shaochen Zhong, Bing Yin, and Xia Hu. 2024.
\newblock \href {https://doi.org/10.1145/3649506} {{Harnessing the Power of LLMs in Practice: A Survey on ChatGPT and Beyond}}.
\newblock \emph{ACM Trans. Knowl. Discov. Data}.

\bibitem[{Zerva et~al.(2022)Zerva, Blain, Rei, Lertvittayakumjorn, C.~de Souza, Eger, Kanojia, Alves, Or{\u{a}}san, Fomicheva, Martins, and Specia}]{zerva-etal-2022-findings}
Chrysoula Zerva, Fr{\'e}d{\'e}ric Blain, Ricardo Rei, Piyawat Lertvittayakumjorn, Jos{\'e}~G. C.~de Souza, Steffen Eger, Diptesh Kanojia, Duarte Alves, Constantin Or{\u{a}}san, Marina Fomicheva, Andr{\'e} F.~T. Martins, and Lucia Specia. 2022.
\newblock \href {https://aclanthology.org/2022.wmt-1.3} {Findings of the {WMT} 2022 shared task on quality estimation}.
\newblock In \emph{Proceedings of the Seventh Conference on Machine Translation (WMT)}, pages 69--99, Abu Dhabi, United Arab Emirates (Hybrid). Association for Computational Linguistics.

\bibitem[{Zheng et~al.(2024)Zheng, Zhang, Zhang, Ye, Luo, Feng, and Ma}]{zheng2024llamafactory}
Yaowei Zheng, Richong Zhang, Junhao Zhang, Yanhan Ye, Zheyan Luo, Zhangchi Feng, and Yongqiang Ma. 2024.
\newblock \href {http://arxiv.org/abs/2403.13372} {{LlamaFactory: Unified Efficient Fine-Tuning of 100+ Language Models}}.
\newblock In \emph{{Proceedings of the 62nd Annual Meeting of the Association for Computational Linguistics (Volume 3: System Demonstrations)}}, Bangkok, Thailand. Association for Computational Linguistics.

\end{thebibliography}

\end{document}